\documentclass{article}
\usepackage{iclr2026_conference,times}

\usepackage[utf8]{inputenc} % allow utf-8 input
\usepackage[T1]{fontenc}    % use 8-bit T1 fonts
\usepackage{url}            % simple URL typesetting
\usepackage{booktabs}       % professional-quality tables
\usepackage{amsfonts}       % blackboard math symbols
\usepackage{nicefrac}       % compact symbols for 1/2, etc.
\usepackage{microtype}      % microtypography
\usepackage{xcolor}         % colors
\usepackage{amssymb}
\usepackage{bbm}
\usepackage{bm}
\usepackage{amsmath}
\usepackage{array,longtable}
\usepackage{caption}
\usepackage{enumitem}
\setlist[itemize]{leftmargin=*}
\setlist[enumerate]{leftmargin=*}
\setlist[description]{leftmargin=*}
\usepackage{longtable,booktabs,array}
\usepackage{xcolor} % in
\usepackage{bbm}

\usepackage{tikz}
\usetikzlibrary{shapes.geometric}
\usepackage{pgfplots}
\pgfplotsset{compat=1.18}
\usepackage{tcolorbox}
\tcbuselibrary{skins,raster}
\usepackage{tikz}
\usetikzlibrary{tikzmark,calc,arrows.meta}

\usepackage{multirow}
\usepackage{graphicx}
\usepackage{tabularx}
\usepackage{algorithm}
\usepackage{algorithmic}
\usepackage{float}
\usepackage{enumitem}

\usepackage{hyperref}       % hyperlinks (load late for float/link patches)
\usepackage{cleveref}       % must be loaded after hyperref + amsmath
\usepackage{amsthm}

\definecolor{paletteEvolve}{HTML}{1748BA}
\definecolor{paletteBestOfN}{HTML}{A8201A}

\usepackage{listings}
\newcommand{\pa}{\textsuperscript{*}} % primary author
\newcommand{\sa}{\textsuperscript{+}} % senior author

\title{Duel-Evolve: Reward-Free Test-Time Scaling via LLM Self-Preferences}

\author{%
  Sweta Karlekar\pa \\
  Columbia University\\
  \And
  Carolina Zheng\pa \\
  Columbia University \\
  \And
  Magnus Saebo\pa \\
  Columbia University \\
  \And
  Nicolas Beltran-Velez \\
  Columbia University \\
  \And
  Shuyang Yu \\
  Columbia University \\
  \And
  John Bowlan\sa \\
  Los Alamos National Laboratory \\
  \And
  Michal Kucer\sa \\
  Los Alamos National Laboratory \\
  \And
  David Blei\sa \\
  Columbia University \\
}

\iclrfinalcopy % Uncomment for camera-ready version, but NOT for submission.

\begin{document}

\maketitle

\begingroup
\renewcommand\thefootnote{}
\footnotetext{\pa~Primary authors, equal contribution. \sa~Senior authors.}
\addtocounter{footnote}{-1}
\endgroup

\begin{abstract}

Many applications seek to optimize LLM outputs at test time by iteratively proposing, scoring, and refining candidates over a discrete output space. Existing methods use a calibrated scalar evaluator for the target objective to guide search, but for many tasks such scores are unavailable, too sparse, or unreliable.
Pairwise comparisons, by contrast, are often easier to elicit, still provide useful signal on improvement directions, and can be obtained from the LLM itself without external supervision.
Building on this observation, we introduce \textsc{Duel-Evolve}, an evolutionary optimization algorithm that replaces external scalar rewards with pairwise preferences elicited from the same LLM used to generate candidates. 
\textsc{Duel-Evolve} aggregates these noisy candidate comparisons via a Bayesian Bradley--Terry model, yielding uncertainty-aware estimates of candidate quality. These quality estimates guide allocation of the comparison budget toward plausible optima using Double Thompson Sampling, as well as selection of high-quality parents to generate improved candidates.
We evaluate \textsc{Duel-Evolve} on MathBench, where it achieves 20 percentage points higher accuracy over existing methods and baselines, and on LiveCodeBench, where it improves over comparable iterative methods by over 12 percentage points. Notably, the method requires no reward model, no ground-truth labels during search, and no hand-crafted scoring function.
Results show that pairwise self-preferences provide strong optimization signal for test-time improvement over large, discrete output spaces.

\end{abstract}

\section{Introduction}
Problems involving natural or symbolic language naturally lend themselves to optimization over a multidimensional discrete space: we seek a candidate \(y \in \mathcal{Y}\) that attains a high value under an objective \(f:\mathcal{Y}\to\mathbb{R}\).
For example, in MaxSAT, $y$ is a Boolean assignment and $f(y)$ counts satisfied clauses \citep{biere2021handbook,bacchus2019maxsat}; in proof search, $y$ is a candidate proof and $f(y)\in\{0,1\}$ indicates whether the target proposition is established \citep{polu2020gptf,bansal2019holist,yang2023leandojo}. Test-time refinement of an LLM response \citep{madaan2023self, snell2024scaling} also fits this paradigm, where $y$ is a token sequence and $f(y)$ measures solution quality for some downstream task.

While the problem is easy to formulate, optimization over $\mathcal{Y}$ is challenging. The discrete space is combinatorially large, small edits can impact quality discontinuously, and gradients, the signal that drives modern machine learning optimization, are undefined. Recent work addresses these difficulties by using LLMs as semantic-aware optimizers: they maintain a population of candidates, evaluate each by assigning a scalar reward $s$ from a surrogate function approximating $f$, and prompt the LLM with $(y,s)$ pairs to propose improved variants \citep{romera2024mathematical,novikov2025alphaevolve}. When the surrogate is informative, the score plays the role of a noisy gradient signal: it induces a consistent ordering over candidates and provides a dense direction for local improvement.

However, in many domains, an informative real-valued surrogate is unavailable or unreliable. Binary verifiers and stochastic batch scores are too sparse or noisy to provide effective guidance for search. A natural way to remove the requirement of specifying an external surrogate is to prompt the LLM itself to qualify or score its responses. However, such ratings may still require an externally specified rubric, and are often poorly calibrated and mutually inconsistent \citep{zheng2023judging}. 

In this paper, we propose to use a different signal that remains internal to the model yet is typically more stable: pairwise preference. Concretely, we use the same LLM as both generator and judge. We query it to choose a winner between two candidates $(y_i,y_j)$ and pool these duels to recover a global estimate of quality---an implicit surrogate for $f$ defined without any external feedback.

Using preferences as the sole optimization signal introduces two algorithmic challenges. First, comparisons are local and noisy, so the algorithm must aggregate them into a coherent global ranking while accounting for statistical uncertainty. Second, comparisons are expensive: under a limited evaluation budget, the algorithm must decide which pairs to compare so that effort is concentrated on candidates that are still 
plausible optima, rather than on those known to be suboptimal.

To address both challenges, we propose \textsc{Duel-Evolve}, an evolutionary optimizer for discrete structured spaces that is guided solely by pairwise preferences.
\textsc{Duel-Evolve} maintains a growing pool of candidates and alternates between (i) selecting informative pairs to compare, (ii) fitting a Bayesian Bradley--Terry model~\citep{bradleyterry1952rank} to aggregate all observed outcomes, and (iii) conditioning the generator LLM on a small set of high scoring parents along with their estimated posterior utilities to propose new candidates.

\textsc{Duel-Evolve} efficiently explores the space of possible solutions through a Bayesian model of global scores, while exploiting the LLM's understanding of language to produce high quality proposals. We use a Laplace approximation to obtain per-candidate posterior means and confidence intervals, and we adapt Double Thompson Sampling (DTS) \citep{wu2016double} to allocate the comparison budget
toward competitive candidates. 

We demonstrate \textsc{Duel-Evolve} on mathematical reasoning and code generation tasks. On MathBench~\citep{liu2024mathbench}, our method reaches 94\% accuracy, exceeding the strongest baseline by 20 percentage points; for LiveCodeBench~\citep{jain2024livecodebench}, it achieves 37\% accuracy, improving over similar evolutionary methods by over 12 percentage points. Notably, these gains are obtained without training a reward model or designing a task-specific scalar surrogate, as the LLM generates both the proposals and comparative judgments. 

\section{Method}
\label{sec:method}

We develop \textsc{Duel-Evolve}, an LLM-based optimization algorithm for
discrete structured spaces guided only by pairwise preferences.
We first describe our problem setting (\S\ref{sec:problem_setup}). We then cast it as a dueling-bandits problem and motivate an idealized Bayesian solution (\S\ref{sec:bayesian_model}). Finally, we derive practical approximations that yield \textsc{Duel-Evolve} (\S\ref{sec:dts-to-de}).

\subsection{Problem Setting}
\label{sec:problem_setup}

Given a query $x$, let $\mathcal{Y}$ be a discrete space of candidate solutions
(e.g., programs, proofs, reasoning traces, or prompt variants), and let
$f:\mathcal{Y}\to\mathbb{R}$ denote an unknown latent utility function that implicitly depends on $x$.
Our goal is to find a maximizer
\begin{equation}
\label{eq:objective}
y^* \;=\; \arg\max_{y \in \mathcal{Y}} f(y).
\end{equation}

We consider a setting in which $f$ is
well-defined but never observed during optimization. 
Instead, optimization proceeds solely through an LLM, which we use in two roles.  First, as a
noisy \emph{judge} $\mathcal{J}$ that, given two candidates
$y_i,y_j\in\mathcal{Y}$ for the same query $x$, returns which candidate it
prefers. 

Second, the LLM also acts as a \emph{generator} $p_\phi$ that, given $x$ and
a small set of parent solutions $A \subset \mathcal{Y}$, proposes new candidates
in $\mathcal{Y}$.

The core algorithmic problem is then how to reliably move
toward~$y^*$ using only pairwise preferences and LLM-based generation,
without access to an external reward model or verifier, under a limited
budget of LLM calls.

\subsection{A First Approach: Double Thompson Sampling with Dueling Bandits}
\label{sec:bayesian_model}

\begin{figure}
    \centering
    \includegraphics[width=1\linewidth]{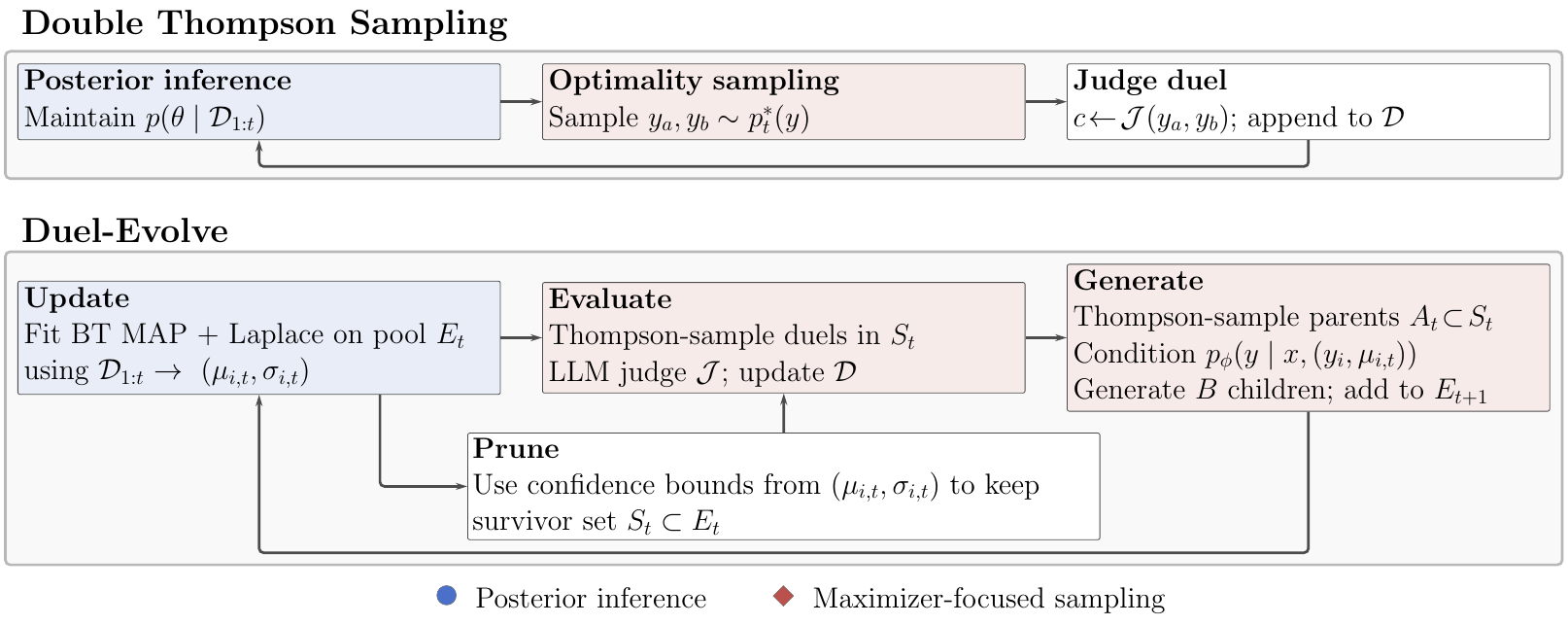}
\caption{\textsc{Duel-Evolve} approximates Double Thompson Sampling (DTS) over a combinatorial space. 
At round $t$, DTS maintains a posterior $p(\boldsymbol{\theta}\mid D_{1:t})$
over latent utilities $\boldsymbol{\theta}=(\theta_y)_{y\in\mathcal{Y}}$
given comparison history $D_{1:t}=\{(y_i,y_j,c_{ij})\}$, and selects the
next duel by sampling $y_a,y_b\sim p_t^*(y)=P\!\big(y=\arg\max_{y'}\theta_{y'}
\mid D_{1:t}\big)$.
\textsc{Duel-Evolve} approximates
\textcolor{paletteEvolve}{posterior inference} by fitting a Bradley--Terry
model on the evaluated pool $E_t$ with a Laplace approximation, yielding
per-candidate summaries $(\mu_{i,t},\sigma_{i,t})$ (\S\ref{sec:operation-i}); it then approximates \textcolor{paletteBestOfN}{maximizer-focused sampling} by
Thompson-sampling duels and parents $A_t$ from a pruned survivor set
$S_t\subseteq E_t$, and proposing children via a conditioned LLM generator
$y\sim p_\phi\!\big(y\mid x,\{(y_i,\mu_{i,t})\}_{y_i\in A_t}\big)$
(\S\ref{sec:operation-ii}--\S\ref{sec:algorithm-overview}).
}
    \label{fig:alg-flow}
\end{figure}

With access only to pairwise comparisons, we adopt the \emph{dueling bandits} framework~\citep{yue2012k} as a convenient formulation of Problem~\eqref{eq:objective}. In this framework, we are given a set of arms $\mathcal{Y}$, and optimization proceeds iteratively: at each round, the algorithm selects a pair of arms $(y_i, y_j)$ to compare, observes noisy binary feedback indicating which is preferred, and uses this outcome to refine its estimates of arm quality, with the goal of identifying the best arm $y^*$ under a limited budget of comparisons.

The difficulty is that pairwise comparisons are both \emph{local} --- each
involves only two candidates --- and \emph{noisy}. Identifying the best arm
therefore requires (i) aggregating many such noisy local signals into a global
ranking of the arms while (ii) deciding which pairs to compare next under a limited budget.

We address both by assigning a latent utility to each arm and modeling
comparisons as noisy functions of utility differences. This yields estimates of the quality of the arms, quantifies remaining
uncertainty, and guides the choice of the next pair to compare. Below, we formalize this idea with a Bradley--Terry likelihood and describe Double Thompson
Sampling (DTS)~\citep{wu2016double}, an idealized algorithm for this setting. We then discuss the obstacles that prevent the
direct application of DTS, motivating the approximations that lead to our algorithm
in~\S\ref{sec:algorithm-overview}.

\paragraph{Bradley--Terry Model.}
Let $\boldsymbol{\theta} = (\theta_y)_{y \in \mathcal{Y}} \in
\mathbb{R}^{|\mathcal{Y}|}$ be a vector of latent utilities, one per
candidate, so that a higher $\theta_y$ indicates a better solution~$y$.
We place a prior $\Pi(\boldsymbol{\theta})$ over this vector to encode
structural assumptions about $\mathcal{Y}$ --- for instance, that
semantically similar solutions should have similar utilities.

Given a pair $(y_i, y_j)$, we model the judge's feedback with a
Bradley--Terry model~\citep{bradleyterry1952rank}. It models the
probability that $y_i$ is preferred over $y_j$ as
\begin{equation}
  P(c_{ij} = +1) = \sigma\,\!\big(\theta_i - \theta_j\big),
  \label{eq:comparison_model}
\end{equation}
where $\sigma$ denotes the logistic sigmoid and
$c_{ij} \in \{-1, +1\}$.

Given all comparisons observed up to round $t$, denoted $\mathcal{D}_{t} = \{(y_i, y_j, c_{ij})\}$, we combine the prior $\Pi(\boldsymbol{\theta})$ with the Bradley--Terry likelihood to obtain the posterior $p(\boldsymbol{\theta} \mid \mathcal{D}_{t})$. This posterior aggregates many pairwise comparisons into a global distribution over the utility of each arm.

\paragraph{Algorithm: Double Thompson Sampling.}
Given the posterior, the algorithm must decide which pair to compare next.
Intuitively, comparisons are most informative when they involve candidates
that are plausibly optimal. Spending comparisons on
arms that the posterior already deems suboptimal with high confidence yields
little new information.

This intuition is formalized by the posterior probability that a given candidate is optimal, its  \emph{maximizer probability},
\begin{equation}
     \label{eq:maximizer_posterior}
  p_t^*(y) \;\equiv\; P\!\big(y = y^* \mid \mathcal{D}_{t}\big) = \int \mathbbm{1}\!\Big[y = \textstyle\arg\max_{y'} \theta_{y'}\Big]\; p(\boldsymbol{\theta} \mid \mathcal{D}_{t})\, d\boldsymbol{\theta}.
\end{equation}
Double Thompson Sampling~\citep{wu2016double} uses $p_t^*$ to select
which pair to compare: at each round, independently draw
$y_a, y_b \sim p_t^*(\cdot)$, query the judge, and update Bradley--Terry posterior. After $T$ rounds, it returns
$\hat{y} = \arg\max_{y} \mu_{y,T}$, where $\mu_{y,T}$ is the
posterior mean utility.
The procedure is summarized in the top half of \Cref{fig:alg-flow}.

\subsection{From Double Thompson Sampling to \textsc{Duel-Evolve}}
\label{sec:dts-to-de}

\textsc{Duel-Evolve} follows a similar structure to Double Thompson Sampling. It maintains a pool of candidates, selects pairs to compare, observes which candidate the judge prefers, and updates a posterior over latent utilities. This posterior then guides the next comparison.
\textsc{Duel-Evolve} extends this loop by also proposing new, improved candidates via LLM-based generation at each iteration. 

However, because $\mathcal{Y}$ is a combinatorially large space of natural-language solutions, two operations that are tractable in the classical finite-arm setting must now be approximated: (i)~maintaining and sampling from the posterior $p(\boldsymbol{\theta} \mid \mathcal{D}_{t})$ and (ii)~sampling candidates proportional to their posterior probability of being optimal, $p_t^*(y)$. We address each in turn (\S\ref{sec:operation-i}--\S\ref{sec:operation-ii}), then show how the two approximations combine to form the \textsc{Duel-Evolve} loop (\S\ref{sec:algorithm-overview}).

\subsubsection{Operation (i): Approximating $p(\boldsymbol{\theta} \mid \mathcal{D}_{t})$}
\label{sec:operation-i}

To approximate the first operation $p(\boldsymbol{\theta} \mid \mathcal{D}_{t})$, rather than defining the posterior over all of $\mathcal{Y}$, we restrict inference to the finite set of candidates that the algorithm has generated and compared so far (how candidates enter this set is described in \S\ref{sec:algorithm-overview}). We assign a latent utility $\theta_i$ to each such candidate $y_i$ and place an independent Gaussian prior $\theta \sim \mathcal{N}(0,\, \sigma_0^2 I)$.

The posterior is unimodal under this prior, so we compute the MAP estimate
\begin{equation}
\label{eq:map}
\hat{\boldsymbol{\theta}}_t
= \arg\max_{\theta}
\left[
  \sum_{(y_i,\, y_j,\, c)\,\in\, \mathcal{D}_{t}}
    \log \sigma\!\bigl(c(\theta_i - \theta_j)\bigr)
  \;-\;
  \frac{\|\theta\|^2}{2\sigma_0^2}
\right],
\end{equation}
where $c \in \{-1,+1\}$. The regularizer resolves the additive non-identifiability inherent to pairwise models. 
We use a Laplace approximation around $\hat{\theta}_t$ with a diagonal Hessian. This yields per-candidate Gaussian summaries $(\mu_{i,t},\, \sigma_{i,t}^2)$ which we can use to judge the quality of each candidate and our uncertainty about it. 
In practice, we solve this problem using an L-BFGS solver
\citep{liu1989limited}.

\subsubsection{Operation (ii): Approximating \texorpdfstring{$p_t^*$}{p*}}
\label{sec:operation-ii}

Computing $p_t^*(y) = P(y = y^* \mid \mathcal{D}_{t})$ exactly would require drawing $\boldsymbol{\theta} \sim p(\boldsymbol{\theta} \mid \mathcal{D}_{t})$ and solving $\arg\max_{y' \in \mathcal{Y}} \theta_{y'}$, which is intractable over the full solution space. We replace this with an LLM-conditioned proposal: given a set of existing solutions annotated with their estimated qualities, the LLM can infer the relationship between solution structure and score and extrapolate toward improvements. In-context learning thus plays the role that a structured prior over $\mathcal{Y}$ would have played in an exact Bayesian treatment, providing the structural generalization needed to propose plausibly optimal candidates without enumeration.

Concretely, we condition the generator on a parent set $A_t$ together with their posterior means,
\begin{equation}
\label{eq:proposal}
y_{\mathrm{new}}
\;\sim\;
p_\phi\!\left(
  y \;\middle|\; x,\;
  \bigl\{(y_i,\, \mu_{i,t})\bigr\}_{y_i \in A_t}
\right),
\end{equation}
where $A_t$ is selected from the set of candidates that have already been generated and compared.  The exact construction of $A_t$ is detailed in the next section as part of the full algorithm.

%% ---------- Algorithm overview ----------
\subsubsection{Putting it together: the \textsc{Duel-Evolve} loop}
\label{sec:algorithm-overview}

\textsc{Duel-Evolve} combines the two approximations above in an evolutionary loop. The algorithm maintains an evaluated pool $E_t$ of all candidates generated so far, initialized by sampling $N_0$ candidates from $p_\phi(\cdot \mid x)$ conditioned only on the query, with an empty comparison history $\mathcal{D} = \emptyset$. Each generation $t$ then proceeds in three phases in a loop:
\begin{enumerate}
    \item \textbf{Update.} Fit the posterior over $E_t$ using full comparison history $\mathcal{D}_{t}$, obtaining $(\mu_{i,t},\, \sigma_{i,t}^2)$ via~\Cref{eq:map} and the Laplace approximation.

    \item \textbf{Evaluate.} Select a batch of pairs from $E_t$ via Thompson sampling---drawing $\tilde{\theta}_i \sim \mathcal{N}(\mu_{i,t},\, \sigma_{i,t}^2)$ and pairing the top-ranked candidates---then query the judge $J$ on all pairs and append the outcomes to $\mathcal{D}_{t}$.

    \item \textbf{Evolve.} Select parents $A_t$ from $E_t$ via Thompson sampling, sample a batch of $B$ new candidates from $p_\phi(\cdot \mid x, A_t)$ via~\Cref{eq:proposal}, and add them to $E_t$.
\end{enumerate}
Within each phase, all judge queries (Step~2) and all generation calls (Step~3) are issued in parallel, so the wall-clock cost per generation is dominated by a single round of LLM inference rather than by the number of comparisons or children produced. The cycle repeats until the budget is exhausted and in the end we return as output $\hat{y} = \arg\max_{y_i \in E_T} \mu_{i,T}$.

While in theory Thompson Sampling effectively balances the exploration/exploitation tradeoff,  in practice we invoke two additional mechanisms to compensate for the fact that our approximations are not exact. First, we maintain a survivor set $S_t \subseteq E_t$ and prune any candidate whose upper confidence bound falls below the best lower confidence
bound, where the width of these bounds is a hyperparameter. Steps~2 and~3 then operate over $S_t$ rather than $E_t$, avoiding wasted comparisons on confidently suboptimal candidates while retaining them in $E_t$ so their data still informs the posterior. Second, we construct $A_t$ by mixing Thompson-sampled top scorers with recently generated candidates that still have high uncertainty, ensuring new arrivals are always evaluated in the next iteration. \Cref{fig:alg-flow} gives the complete procedure.

\section{Related Work}
\paragraph{Optimization in discrete spaces.} Although our work is applied in the context of LLM-generated outputs, it belongs to the broader tradition of heuristic optimization over discrete and combinatorial spaces, where exact methods are intractable and gradient information is unavailable. Classic metaheuristics such as genetic algorithms \citep{golberg1989genetic} and simulated annealing \citep{kirkpatrick1983optimization} iteratively refine candidate solutions through stochastic perturbation and selection, guided by a scalar objective. Recently, LLMs have been embedded into such search loops—proposing improved candidates conditioned on scored histories \citep{yang2023large}, maintaining evolutionary populations of programs scored by explicit evaluators \citep{romera2024mathematical, novikov2025alphaevolve}, or optimizing prompts for downstream tasks, like GEPA \citep{agrawal2025gepa}, among others \citep{fernando2023promptbreeder, guo2023connecting}. However, all of these approaches rely on an informative scalar evaluator, which is often unavailable or prohibitively expensive to design for open-ended generation tasks. \textsc{Duel-Evolve} removes this requirement by replacing scalar scores with pairwise preferences aggregated through a Bayesian Bradley--Terry model into a global posterior over candidate quality.
Concurrent work, Feedback Descent \citep{lee2025feedback}, also eschews scalar evaluation but performs single-trajectory hill-climbing against a single incumbent, maintaining no global model of candidate quality and offering no mechanism for allocating comparisons efficiently. \textsc{Duel-Evolve} addresses both limitations by fitting a posterior over observed duels to obtain uncertainty-aware quality estimates and using Thompson sampling to focus comparisons on strong candidates. 
Prompt Duel Optimizer (PDO) \citep{wu2025llm}, another concurrent method that forgoes scalar evaluation, applies Double Thompson Sampling with LLM-judged pairwise preferences similarly to \textsc{Duel-Evolve}. However, PDO targets dataset-level prompt selection rather than per-instance response optimization, and uses independent Beta posteriors with a Copeland-style objective rather than our global Bayesian Bradley–Terry posterior over candidate utilities.

\paragraph{Dueling bandits.} \textsc{Duel-Evolve} builds on the dueling-bandits framework, first introduced by \citet{yue2009interactively} for interactive information retrieval and online ranker optimization, and later formalized as the $K$-armed dueling-bandits problem by \citet{yue2012k}. Since then, dueling bandits have become a general framework for preference-based online learning and comparison-based selection \citep{ailon2014reducing, bengs2021preference}. Standard formulations assume a fixed, finite arm set and a winner notion such as a Condorcet winner, and design sampling rules to minimize regret or identify the best arm. Prominent algorithms include frequentist confidence-bound methods \citep{yue2012k, zoghi2014relative, zoghi2015mergerucb, komiyama2015regret} and Bayesian posterior-sampling rules such as DTS \citep{wu2016double}, and are often paired with parametric preference models like Bradley--Terry or Plackett--Luce \citep{saha2021optimal}. While \textsc{Duel-Evolve} adopts a similar parametric formulation and is inspired by standard algorithms such as DTS, it diverges from these works by operating over a combinatorially large arm set that grows during optimization, requiring approximate inference and pruning heuristics in place of the exact computations of the finite-arm literature.

\textbf{Test-time search, refinement, and compute scaling for LLM outputs.} \textsc{Duel-Evolve} is also part of the growing family of methods that improve LLM outputs by scaling test-time computation~\citep{snell2024scaling}. These methods can be organized along two axes: whether they require an external reward signal, and whether performance continues to improve as more compute is invested. Methods relying on external evaluators---Best-of-$N$ with trained verifiers~\citep{cobbe2021training}, process reward models with step-level tree search~\citep{lightman2023let}, and repeated sampling at scale~\citep{brown2024large}---are often bottlenecked by scorer quality and availability. Methods that avoid external reward---reasoning models~\citep{openai2024o1, guo2025deepseek}, iterative self-refinement~\citep{madaan2023self, shinn2023reflexion}, and search over intermediate steps~\citep{yao2023tree}---typically operate on a single candidate with no mechanism to pool information across solutions and have an upper bound on scaling. \textsc{Duel-Evolve} sits in the quadrant that requires no external reward yet continues improving with additional compute, as more generations yield both better utility estimates and exploration of stronger regions of the solution space. It also complements single-generation methods: for example, reasoning models and refinement techniques can serve as subroutines (e.g., as a stronger judge  or generator) within the algorithm.
 
\section{Experiments}
\label{sec:experiments}

We evaluate \textsc{Duel-Evolve} on two benchmarks: mathematical reasoning (MathBench) and code generation (LiveCodeBench).
Our experiments compare against chain-of-thought prompting, multiple sampling with aggregation, and iterative-refinement baselines. We also examine how performance scales with the number of evolutionary generations.

\subsection{Tasks and Metrics}
\label{sec:tasks}

\paragraph{MathBench.}
MathBench~\citep{liu2024mathbench} is a dataset that contains multiple-choice mathematics questions spanning primary school to college-level curricula. A math problem serves as the query $x$, and the candidate space $\mathcal{Y}$ is the model's reasoning trace paired with a multiple-choice answer. \textsc{Duel-Evolve} evolves these trace-answer pairs.
We use the English-language splits at the \textsc{Middle}, \textsc{High}, and \textsc{College} difficulty levels, restricting to the single-choice format in which each question is presented as a four-option problem with one correct answer. We exclude all ``knowledge'' subsets, which primarily test recall of definitions or formulae, and retain only word-problem-style questions that require multi-step numerical and algebraic reasoning. From the selected subsets, we construct a balanced test set of 150 problems by stratified sampling, with 50 problems per difficulty level. A separate validation split is held out for constructing few-shot demonstrations (see~\S\ref{sec:baselines}). We report accuracy, or the fraction of problems with the correct multiple-choice answer, both over all problems and stratified by difficulty level.
All MathBench experiments use Gemma-3-4b-it \citep{team2024gemma}, a model well-suited to these problem difficulties.

\paragraph{LiveCodeBench.}
LiveCodeBench v6~\citep{jain2024livecodebench} is an execution-based dataset of Python competitive programming problems sourced from \cite{atcoder}, \cite{leetcode}, and \cite{codeforces}.
A programming problem serves as the query $x$, and the candidate space $\mathcal{Y}$ consists of code solutions. Every problem includes a small set of public tests (1--4 per problem; median 3) and a substantially larger hidden test set (up to 100; median 25). We use the public tests only as a coarse feedback signal for all iterative methods during evolution
and report accuracy as the percentage of problems for which 100\% of the hidden test cases pass, following standard practice for LiveCodeBench.
To enable rapid iteration, we construct an evaluation set of 99 problems, selecting 33 problems from each difficulty level: \textsc{Easy}, \textsc{Medium}, and \textsc{Hard}.
Because the competitive programming problems in LiveCodeBench require stronger reasoning skills than those in MathBench, we use the larger Gemma-3-27b-it \citep{team2024gemma} (4-bit quantized; \cite{redhatai_gemma3_27b_it_quantized_w4a16_2025}) for all LiveCodeBench experiments.

The two tasks studied exemplify the sparse-reward settings that motivate our approach.
On MathBench, the only signal is final correctness, with no intermediate feedback to guide search.
On LiveCodeBench, each problem exposes 1--4 public tests, which provide a coarse intermediate reward. However, a solution that passes all public tests may still fail on the hidden suite, so optimizing against public tests alone is insufficient.
In both tasks, the absence of a rich scalar objective reduces the effectiveness of score-based search methods, motivating the use of pairwise preferences as an alternative signal for guiding evolutionary search.

\subsection{Baselines}
\label{sec:baselines}
We compare \textsc{Duel-Evolve} against baselines that operate under increasingly privileged information regimes.
Problem-only methods receive only the problem statement: zero-shot chain-of-thought~\citep{wei2022chain}, self-consistency~\citep{wang2022self}, Feedback Descent~\citep{lee2025feedback}, Best-of-$N$---which selects from $N$ i.i.d.\ samples via the same pairwise preference mechanism used by \textsc{Duel-Evolve} but without the evolutionary loop---and \textsc{Duel-Evolve} itself.
Demonstration-aided methods additionally condition on solved examples: few-shot chain-of-thought prepends $k$ exemplars with reasoning traces from a held-out validation split ($k{=}8$ for MathBench and $k{=}3$ for LiveCodeBench).
Label-aided methods further have access to per-instance correctness: we evaluate GEPA~\citep{agrawal2025gepa}, which uses ground-truth supervision to maintain a Pareto frontier and drive reflective mutation of a system prompt used to generate problem solutions.
All methods use the same underlying language model; further baseline details are provided in Appendix~\ref{app:baselines}.

\subsection{Main Results}
\label{sec:main_results}

\begin{figure}[t]
\centering
\begin{minipage}[t]{0.66\linewidth}
  \vspace{0pt}
  \centering
  \includegraphics[width=\linewidth]{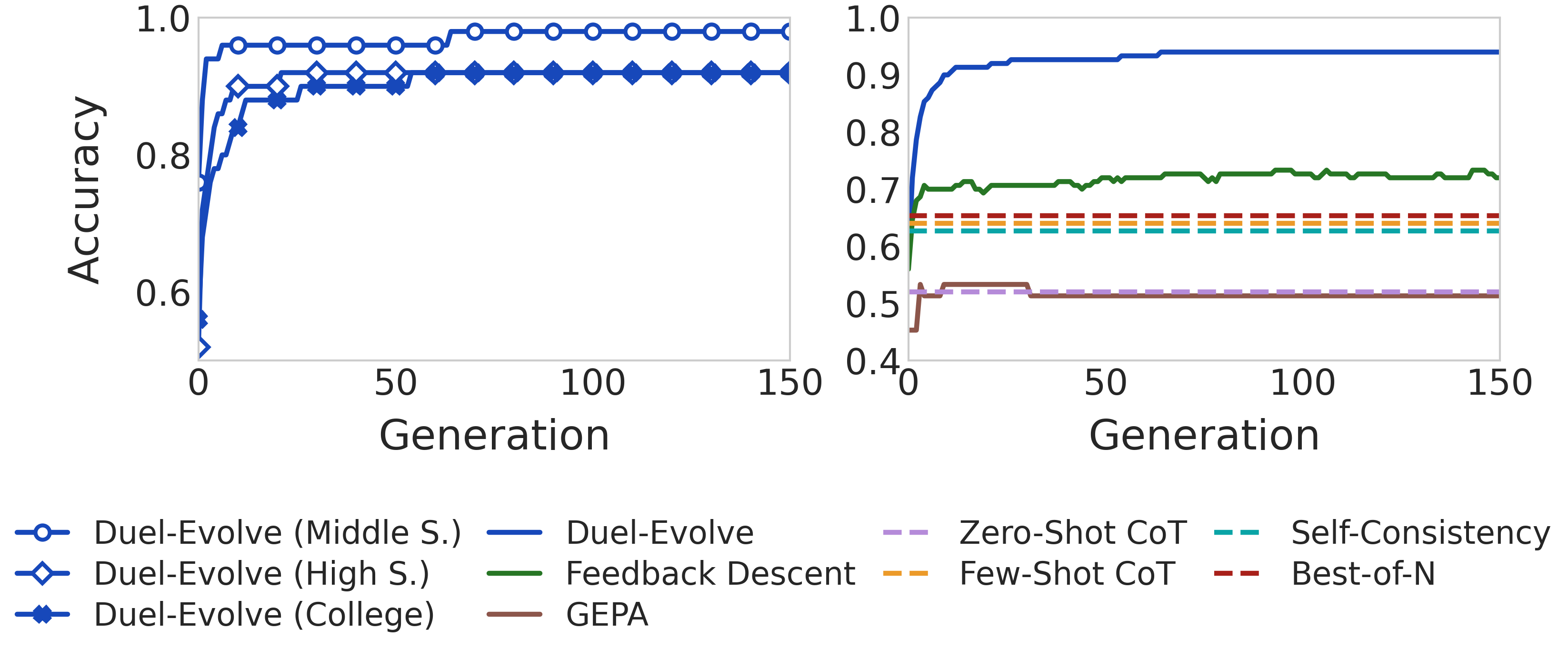}
\end{minipage}\hfill
\begin{minipage}[t]{0.34\linewidth}
  \vspace{0pt}
  \centering
  \renewcommand{\arraystretch}{1.1}
  \setlength{\tabcolsep}{4pt}
  \small
  \begin{tabular}{lc}
    \toprule
    Method & Accuracy (\%) \\
    \midrule
    Zero-shot CoT & 57.3 \\
    Few-shot CoT ($k{=}8$) & 64.0 \\
    Self-consistency & 62.7 \\
    Best-of-$N$ & 65.3 \\
    GEPA & 51.3 \\
    Feedback descent & 72.0 \\
    \midrule
    \textsc{Duel-Evolve} & \textbf{94.0} \\
    \bottomrule
  \end{tabular}
\end{minipage}
\caption{\textbf{MathBench accuracy over 150 generations.} \textbf{Left:} \textsc{Duel-Evolve} performance stratified by difficulty level (\textsc{Middle}, \textsc{High}, \textsc{College}). \textbf{Middle:} Method comparison over generations: non-iterative baselines (Zero-shot CoT, Few-shot CoT, Self-consistency, and Best-of-$N$) remain flat, while iterative methods (Feedback Descent, GEPA, and \textsc{Duel-Evolve}) improve over time. \textbf{Right:} Final accuracy across methods. \textsc{Duel-Evolve} achieves the best performance.}
\label{fig:mathbench}
\end{figure}

We report accuracy across all methods for both MathBench and LiveCodeBench.
For iterative methods, we report the accuracy of the identified best solution at each generation step. 

The table in \Cref{fig:mathbench} reports final accuracy for each method on MathBench after 150 generations.
\textsc{Duel-Evolve} achieves 94\% accuracy, exceeding the strongest baseline by 22 percentage points.
Sampling methods perform modestly, with self-consistency, Best-of-$N$, and few-shot prompting all achieving similar accuracies.
GEPA did not perform as well---we believe this is because the optimized prompts overfit to the validation set, on which the best prompt achieved 67\% accuracy. Iterative refinement via Feedback Descent improves over static methods and GEPA, but \textsc{Duel-Evolve} achieves the highest accuracy given the iteration budget. 
Its accuracy as a function of generation step (\Cref{fig:mathbench}, left and middle) shows rapid early improvement, rising from 57\% to 90\% within the first 10 generations, followed by slower improvement to 94\% by generation~64.
The difficulty-stratified curves (left) show that \textsc{middle} problems are solved fastest, reaching 96\% by generation~6 and 98\% at convergence.
\textsc{High} and \textsc{college} problems follow a similar trajectory but converge more slowly, both reaching 92\%.

\begin{figure}[t]
\centering
\begin{minipage}[t]{0.66\linewidth}
  \vspace{0pt}
  \centering
  \includegraphics[width=\linewidth]{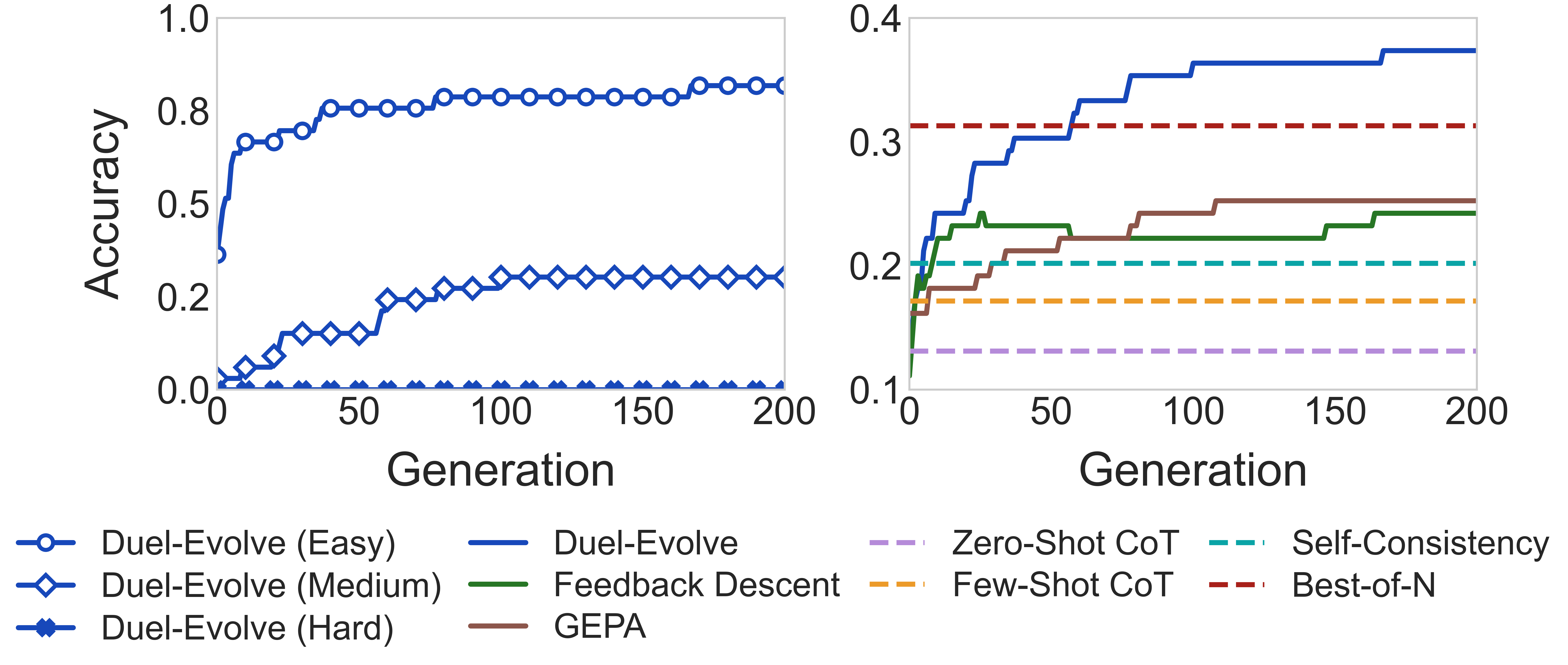}
\end{minipage}\hfill
\begin{minipage}[t]{0.34\linewidth}
  \vspace{0pt}
  \centering
  \renewcommand{\arraystretch}{1.11}
  \setlength{\tabcolsep}{4pt}
  \small
  \begin{tabular}{lc}
    \toprule
    Method & Accuracy (\%) \\
    \midrule
    Zero-shot CoT & 13.1 \\
    Few-shot CoT ($k{=}3$) & 16.2 \\
    Self-consistency & 20.2 \\
    Best-of-$N$ & 31.3 \\
    GEPA & 25.3 \\
    Feedback Descent & 24.2 \\
    \midrule
    \textsc{Duel-Evolve} & \textbf{37.4} \\
    \bottomrule
  \end{tabular}
\end{minipage}
\caption{\textbf{LiveCodeBench accuracy over 200 generations.}
\textbf{Left:} \textsc{Duel-Evolve} performance stratified by difficulty level (\textsc{Easy}, \textsc{Medium}, \textsc{Hard}). \textbf{Middle:} Method comparison over generations: static baselines (Zero-shot CoT, Few-shot CoT, Self-consistency, and Best-of-$N$) are flat, while iterative methods (Feedback Descent, GEPA and \textsc{Duel-Evolve}) improve over time. \textbf{Right:} Final accuracy across methods. \textsc{Duel-Evolve} achieves the best performance.}

\label{fig:lcb}
\end{figure}

\Cref{fig:lcb} reports results on LiveCodeBench after 200 generations. As on MathBench, \textsc{Duel-Evolve} outperforms the baselines, attaining 37.4\% accuracy and exceeding Feedback Descent and GEPA by over 12 percentage points. Notably, \textsc{Duel-Evolve} surpasses other iterative baselines by the fifth generation (\Cref{fig:lcb}, middle). Best-of-$N$ performs competitively, suggesting that our preference-based selection procedure provides a meaningful signal even without iterative search; \textsc{Duel-Evolve} builds on this to achieve further gains. In the difficulty-stratified accuracy plot, we see large early improvement followed by steady accuracy gains in both the \textsc{Easy} and \textsc{Medium} problems. 
Unlike in our MathBench experiments, GEPA outperforms the sampling-based baselines, except for Best-of-$N$, and even exceeds the final accuracy of Feedback Descent. However, due to the substantially higher wall-clock cost of the reference GEPA implementation on LiveCodeBench problems, we limited GEPA to 125 iterations and extrapolate its accuracy at iteration 125 for (middle) in \Cref{fig:lcb}. We made a best-faith effort to optimize its throughput in our setup, but the method remained significantly more expensive than other baselines.

\section{Discussion}

\textsc{Duel-Evolve} demonstrates that pairwise preferences elicited from an LLM can serve as effective optimization signals when informative scalar rewards are unavailable. Choosing a winner between two candidates is typically an easier task for an LLM than producing an optimal solution directly. \textsc{Duel-Evolve} exploits this by aggregating many noisy local duels into global quality estimates via the Bayesian Bradley–Terry model, and using the model posteriors to generate proposals from stronger regions of the solution space. Notably, \textsc{Duel-Evolve} does not use an external reward model, ground-truth labels, or task-specific scoring functions. On MathBench, where the only feedback is final binary correctness, \textsc{Duel-Evolve} reaches 94.0\% accuracy, exceeding the strongest baseline by over 20 percentage points. On LiveCodeBench, where public test coverage is insufficient to guarantee hidden-test success, it attains 37.4\% accuracy, improving over other iterative methods by over 12 percentage points.

It is important to note that, relative to $k$-shot baselines, all iterative methods incur additional inference cost from repeated generation and judging. Nevertheless, on MathBench, approximately 90\% of the total improvement occurs within the first 10 generations, demonstrating rapid convergence is possible depending on the task. Further, on both benchmarks, \textsc{Duel-Evolve} performance separates from the other iterative methods early in the search, with the rate of improvement exceeding that of Feedback Descent and GEPA within the first few generations.

Because the optimization signal of \textsc{Duel-Evolve} derives entirely from the model's own preferences, a limitation of the method is that it will amplify rather than correct systematic judge biases, such as a preference for confidence over correctness. This may be especially apparent in more open-ended domains such as summarization, dialogue, or creative generation, where quality criteria are subjective. Mitigating such biases in the context of \textsc{Duel-Evolve} through ensembling models or calibrating against labeled subsets remains important future work.

\bibliography{references}
\bibliographystyle{iclr2026_conference}

\appendix 
\section{Method Details}

\subsection{Implementation Details}
\label{sec:implementation}

All experiments use the same model for both generation and evaluation.
For \textsc{Duel-Evolve} on MathBench, each evolutionary generation produces a batch of 12 candidate solutions, each conditioned on 6 scored parents selected via a Thompson-sampling / recency mixture. 

For \textsc{Duel-Evolve} on LiveCodeBench, each evolutionary generation produces a batch of 40 candidate solutions, each conditioned on 5 scored parents selected via a Thompson-sampling / recency mixture. Additionally, we employ abstract syntax tree-based (AST-based) de-duplication of candidates for LiveCodeBench, where AST-based signatures for each candidate are stored. If a candidate with a repeating signature would be added to the candidate pool, it is instead discarded and not replaced.

Evaluation uses order-consistent pairwise judging: each pair is judged twice with swapped presentation order, and only concordant judgments are retained as decisive comparisons.
The Bradley--Terry posterior (MAP + Laplace) is updated after each evaluation round, and the population is maintained at a maximum of 200 solutions with confidence-based pruning.
Responses are generated in structured JSON format with explicit reasoning and answer fields.

We retain only \emph{order-consistent decisive} outcomes,
\[
d_{ij}=\mathbbm{1}\!\left[(z^{(1)}_{ij}=A \wedge z^{(2)}_{ij}=B)\;\vee\;(z^{(1)}_{ij}=B \wedge z^{(2)}_{ij}=A)\right].
\]
If $d_{ij}=1$, we record a directed preference $y_i\succ y_j$ or $y_j\succ y_i$ accordingly. If $d_{ij}=0$, the comparison is treated as non-informative and excluded from model updates. Operationally, this is equivalent to dropping ties before fitting the Bradley--Terry model.

For solution generation for MathBench, we use a temperature of 0.7. For LiveCodeBench, we increase the temperature to 1.2 as we noticed duplicate solutions. After adding AST-based de-duplication, for some problems half of the generation batch would get de-duplicated. To remedy this, we increased the temperature to promote more diverse generations. This likely explains why Best-of-$N$ performs relatively better on LiveCodeBench than on MathBench, as the higher temperature naturally leads to more exploration. 

For LiveCodeBench, we additionally add ``evolving memory" to the generation prompt. Evolving memory acts as a model scratchpad and is preserved through each iteration in an evolutionary path. At each iteration, the model is allowed to both update the solution and also edit the evolving memory to use as additional short ``notes''. For example, the evolving memory may be used to store key insights for future attempts like edge cases found, algorithmic patterns that work, complexity considerations, etc. We instruct the model to keep the evolving memory to less than 500 characters. 

\textbf{Feedback Descent \citep{lee2025feedback}.} We implemented Feedback Descent based on the paper, as the method does not have a public codebase. To prevent overly long context lengths in the iterative solution generation step, we use a window size of 15 previous solutions.

\textbf{GEPA \citep{agrawal2025gepa}.} We set the base and reflective LLMs to be the same model. For LiveCodeBench, we use GEPA inference-time search, which sets the training and validation splits to be the test set; for MathBench, we use separate training and validation splits, as feedback for an example is identical to its solution. For both datasets, we run GEPA using the AnyMaths Adapter from the GEPA GitHub repository;\footnote{\url{https://github.com/gepa-ai/gepa}} for LiveCodeBench, we modify the adapter to incorporate test case-based evaluation and feedback.

\section{Prompt Templates}
\label{app:prompt-templates}
For all methods except GEPA (\Cref{sec:gepa_prompt}), we use the system prompt "You are a helpful assistant."

\subsection{MathBench}

\subsubsection{Initial Generation Prompt Template (No Parents)}
\begin{lstlisting}
# Problem
{question}

# Instructions
Work through the problem step-by-step, checking each computation or inference
before moving on. Re-review the answer choices as you reason, and only commit
to a final option after verifying it matches your logic. Your final line must
clearly name the single best option letter, justified by the reasoning above it.

# Multiple-Choice Options
{options}

Your response must be JSON with this schema:
{
    "reasoning": "Show your reasoning first. Think step-by-step here. Be concise and straight-forward.",
    "solution": "Based on the reasoning, give your final letter choice answer. E.g., 'A', 'B', 'C', or 'D'."
}
\end{lstlisting}
\par\medskip
\subsubsection{Generation Prompt Template (Evolution / With Parents)}
% \begin{verbatim}
\par\medskip
\begin{lstlisting}

# Problem
{question}

# Options
{options}

# Parent Solutions
{parent_solutions}

# Instructions
The parent solutions are drafts that may include mistakes, gaps, or repetitive
reasoning. Audit them critically: point out any flaws you notice, re-check the
options from scratch, and then rebuild a cleaner line of reasoning that improves
on the parents. Finish by stating the single best option letter that your improved
reasoning supports; never copy a parent answer without verifying it.

Your response must be JSON with this schema:
{
    "reasoning": "Show your reasoning first. Think step-by-step here. Be concise and straight-forward.",
    "solution": "Based on the reasoning, give your final letter choice answer. E.g., 'A', 'B', 'C', or 'D'."
}
\end{lstlisting}
\subsubsection{Template for Each Parent Solution Block ({\ttfamily \{parent\_solutions\}})}
\begin{lstlisting}
    
Answer: {answer_letter} ({option_text_if_available})
Score: {score:.3f}
Reasoning:
{reasoning}
\end{lstlisting}
\par\medskip

\subsubsection{Evaluation Prompt Template (LLM Judge)}
% \begin{verbatim}
\begin{lstlisting}
# Goal
You will be given a math problem and two candidate answers.
Decide which candidate provides the most convincing reasoning and final answer.

# Problem
{question}

# Options
{options}

# Candidate A
{candidate_a}

# Candidate B
{candidate_b}

# Output Format (STRICT)
Your response must be JSON with this schema:
{
    "solution": "Pick A, B, or T (for tie)",
    "reasoning": "Brief explanation of your reasoning (one sentence)"
}
\end{lstlisting}

\subsubsection{Few-Shot Generation Template (k-shot)}
% \begin{verbatim}
\begin{lstlisting}
# Instructions
Work through the problem step-by-step, checking each computation or inference
before moving on. Re-review the answer choices as you reason, and only commit
to a final option after verifying it matches your logic. Your final line must
clearly name the single best option letter, justified by the reasoning above it.

Your response must be JSON with this schema:
{
    "reasoning": "Show your reasoning first. Think step-by-step here. Be  concise and straight-forward.",
    "solution": "Based on the reasoning, give your final letter choice answer. E.g., 'A', 'B', 'C', or 'D'."
}

# Problem
<few_shot_question_1>
# Multiple-Choice Options
A. <option_a_1>
B. <option_b_1>
C. <option_c_1>
D. <option_d_1>
# Response
{"reasoning":"<few_shot_reasoning_1>","solution":"<few_shot_answer_letter_1>"}

# Problem
<few_shot_question_2>
# Multiple-Choice Options
A. <option_a_2>
B. <option_b_2>
C. <option_c_2>
D. <option_d_2>
# Response
{"reasoning":"<few_shot_reasoning_2>","solution":"<few_shot_answer_letter_2>"}

...

# Problem
{question}
# Multiple-Choice Options
{options}
# Response
\end{lstlisting}

\subsubsection{GEPA}
\label{sec:gepa_prompt}
We use a slightly modified prompt for GEPA, as their method optimizes the system prompt.

\textit{System prompt.}
\begin{lstlisting}
You are a helpful assistant. Work through the problem step-by-step, checking each computation or inference before moving on. Re-review the answer choices as you reason, and only commit to a final option after verifying it matches your logic. Your final line must clearly name the single best option letter, justified by the reasoning above it.

Your response must be JSON with this schema:
{
    "reasoning": "Show your reasoning first. Think step-by-step here. Be concise and straight-forward.",
    "solution": "Based on the reasoning, give your final letter choice answer. E.g., 'A', 'B', 'C', or 'D'."
}
\end{lstlisting}

\textit{User prompt.}
\begin{lstlisting}
# Problem
{question}
# Multiple-Choice Options
{options}
\end{lstlisting}

\subsection{LiveCodeBench}

\subsubsection{Initial Generation Prompt Template (No Parents)}
\begin{lstlisting}
Solve the following competitive programming problem.

# Problem
{question}

# Public Examples (I/O)
{options}

{starter_block}

# Instructions
- Your solution MUST be a complete, executable Python program that reads from stdin and prints to stdout.
- Even if a starter code template shows a class, you MUST add code to read input, call the function/method, and print the result.
- Parse input exactly as shown in the Public Examples. If input shows JSON (e.g., `[1, 2, 3]`), use `json.loads(input())`. If input shows space-separated values, use `input().split()`.
- Correctness is the top priority. In your reasoning, trace through your solution on at least one example to verify it produces the correct output.
- Consider edge cases.
- Optimize for time complexity after ensuring correctness.
- Solutions will be evaluated on a hidden test set of inputs with size up to the max of the constraint size mentioned in the problem. Solutions must run within 4 seconds for each test inputs, so check that the time complexity is efficient enough for the largest inputs.
- Max complexity for constraint size N (simpler is okay): N<=12 O(n!), N<=25 O(2^n), N<=500 O(n^3), N<=10^4 O(n^2), N<=10^6 O(n log n), N<=10^8 O(n), N>10^8 O(log n) or O(1).
- Never violate the above constraint on time complexity given constraint size.
- Rule of thumb: ~10^8 operations per second. For N=10^5, O(N^2)=10^10 operations = 100+ seconds (too slow). O(N)=10^5 operations = instant.
- Avoid brute-force algorithms.
- Output only the JSON below.

Your response must be JSON with this schema:
{
    "reasoning": "Brief explanation of your approach and why it should be correct/efficient",
    "evolving_memory": "Key insights for future attempts: edge cases found, algorithmic patterns that work, complexity considerations (max 500 chars)",
    "solution": "```python\n<your complete Python solution>\n```"
}
\end{lstlisting}

\subsubsection{Generation Prompt Template (Evolution / With Parents)}
\begin{lstlisting}
Improve upon previous candidates by considering their faults (edge cases, time complexity, errors in public test execution traces, etc).
The evolving_memory field carries forward observations from previous attempts. Read the parent solutions' evolving_memory to understand what has been tried and what issues were found. In your response, update the evolving_memory with any new insights about the problem.

# Problem
{question}

# Public Examples (I/O)
{options}

{starter_block}

# Parent Solutions (each shows Answer/Reasoning, evolving memory, and public test execution traces)
{parent_solutions}

# Instructions
- Your solution MUST be a complete, executable Python program that reads from stdin and prints to stdout.
- Even if a starter code template shows a class, you MUST add code to read input, call the function/method, and print the result.
- Parse input exactly as shown in the Public Examples. If input shows JSON (e.g., `[1, 2, 3]`), use `json.loads(input())`. If input shows space-separated values, use `input().split()`.
- Never hard-code inputs that are used to print outputs! Always parse an input as above and print the output response based on the parsed input. The examples shown are NOT the hidden test cases used to evaluate solutions.
- Correctness is the top priority. In your reasoning, trace through your solution on at least one example to verify it produces the correct output.
- Consider edge cases.
- Optimize for time complexity after ensuring correctness.
- Solutions will be evaluated on a hidden test set of inputs with size up to the max of the constraint size mentioned in the problem. Solutions must run within 4 seconds for each test inputs, so check that the time complexity is efficient enough for the largest inputs.
- Max complexity for constraint size N (simpler is okay): N<=12 O(n!), N<=25 O(2^n), N<=500 O(n^3), N<=10^4 O(n^2), N<=10^6 O(n log n), N<=10^8 O(n), N>10^8 O(log n) or O(1).
- If parent solutions violate the above constraint on time complexity, it is always better to produce a new, more efficient solution than to copy the inefficient algorithm.
- If parent solutions do not violate the above constraint, always prioritize correctness over efficiency.
- Rule of thumb: ~10^8 operations per second. For N=10^5, O(N^2)=10^10 operations = 100+ seconds (too slow). O(N)=10^5 operations = instant.
- Use insights from evolving_memory to avoid repeating past mistakes.
- Do not hesitate to revert to simpler approaches if "optimized" solutions seem wrong.
- Do not hesitate to make large corrections.
- You MUST make some change. You CANNOT copy verbatim. Try a different algorithm, fix a bug, or optimize the approach.
- Address the issues presented in the public test execution traces. The main purpose of these traces is to spot syntax errors and problems with parsing. Just because all public test cases pass do NOT assume the solutions pass all hidden test cases.
- Avoid brute-force algorithms.
- Output only the JSON below.

Your response must be JSON with this schema:
{
    "reasoning": "Brief explanation of your approach and why it should be correct/efficient",
    "evolving_memory": "Key insights for future attempts: edge cases found, algorithmic patterns that work, complexity considerations (max 500 chars)",
    "solution": "```python\n<your complete Python solution>\n```"
}
\end{lstlisting}

\subsubsection{Template for Each Parent Solution Block ({\ttfamily \{parent\_solutions\}})}
\begin{lstlisting}
Solution:
{parent.text}

Reasoning: {parent.reasoning}

Recent win explanation: {parent.recent_win_explanation}

Recent loss explanation: {parent.recent_loss_explanation}

Public test executions (sample):
- {test_str_1} -> {outcome_str_1}
- {test_str_2} -> {outcome_str_2}
- {test_str_3} -> {outcome_str_3}

Evolving memory: {parent.evolving_memory}
\end{lstlisting}

\subsubsection{Evaluation Prompt Template (LLM Judge)}
\begin{lstlisting}
You are comparing two code solutions for a competitive programming problem.

# Problem
{question}

# Solution A
```python
{code_a}
```

## Execution Results for A
{trace_a}

# Solution B
```python
{code_b}
```

## Execution Results for B
{trace_b}

# Context
Solutions will be evaluated on a hidden test set of inputs with size up to the max of the constraint size mentioned in the problem. Solutions must run within 4 seconds for each test inputs, so check that the time complexity is efficient enough for the largest inputs.                            

# Task
Decide which solution is BETTER. Use the following priority order:

1. EXECUTABILITY: A valid solution MUST be a complete, executable program that:
    - Reads input from stdin (using `input()` or `sys.stdin`)
    - Prints output to stdout (using `print()`)
    - If it defines a class, it MUST also have code outside the class to instantiate and call it
A class-only definition like `class Solution: def solve(...)` without stdin/stdout code is NOT executable.

2. INPUT PARSING COMPATIBILITY: Check if the solution's input parsing matches the problem's input format.
    - A solution should parse input exactly as shown in the Public Examples.
    - If input shows JSON (e.g., `[1, 2, 3]`), use `json.loads(input())`
    - If input shows space-separated values, use `input().split()`
    - If the input parsing doesn't match the problem's format, the solution will produce wrong output and should always LOSE.
                            
3. Time complexity constraint: Max complexity for constraint size N (simpler is okay): N<=12 O(n!), N<=25 O(2^n), N<=500 O(n^3), N<=10^4 O(n^2), N<=10^6 O(n log n), N<=10^8 O(n), N>10^8 O(log n) or O(1). Evaluate the time complexity of the two options, reference the size constraint for the problem, and check against this STRICT complexity constraint.
                            
4. Correctness: Among executable solutions, one that produces correct output beats one that doesn't.

5. Efficiency: Among executable, correct solutions, prefer better time complexity.

6. Edge case handling: Among executable, correct, efficient solutions, prefer those that best handle boundary conditions.

7. Avoid brute-force: Brute-force solutions are likely to be too inefficient.
                            
Consider any failures in the public test execution traces if they exist. The main purpose of these traces is to spot syntax errors and problems with parsing. Just because all public test cases pass, do NOT assume the solutions pass all hidden test cases.
Furthermore, no solution should use hard-coded inputs. Each solution should appropriately parse the input.

Your response must be JSON with this schema:
{
    "reasoning": "Brief explanation of your reasoning (one sentence)",
    "solution": "Pick A, B, or T (for tie)"
}
\end{lstlisting}

\subsubsection{Few-Shot Generation Template (k-shot)}
\begin{lstlisting}
Solve the following competitive programming problem. Work through the problem step-by-step.

# Solved Reference Examples
Below are 3 correctly solved problems for reference. Study the solution patterns.

{few_shot_examples}

---

# Problem to Solve
{question}

# Public Examples (I/O)
{options}

{starter_block}

# Instructions
- Your solution MUST be a complete, executable Python program that reads from stdin and prints to stdout.
- Even if a starter code template shows a class, you MUST add code to read input, call the function/method, and print the result.
- Parse input exactly as shown in the Public Examples. If input shows JSON (e.g., `[1, 2, 3]`), use `json.loads(input())`. If input shows space-separated values, use `input().split()`.
- Correctness is the top priority. In your reasoning, trace through your solution on at least one example to verify it produces the correct output.
- Consider edge cases.
- Optimize for time complexity after ensuring correctness.
- Solutions will be evaluated on a hidden test set of inputs with size up to the max of the constraint size mentioned in the problem. Solutions must run within 4 seconds for each test inputs, so check that the time complexity is efficient enough for the largest inputs.
- Max complexity for constraint size N (simpler is okay): N<=12 O(n!), N<=25 O(2^n), N<=500 O(n^3), N<=10^4 O(n^2), N<=10^6 O(n log n), N<=10^8 O(n), N>10^8 O(log n) or O(1).
- Never violate the above constraint on time complexity given constraint size.
- Rule of thumb: ~10^8 operations per second. For N=10^5, O(N^2)=10^10 operations = 100+ seconds (too slow). O(N)=10^5 operations = instant.
- Avoid brute-force algorithms.
- Output only the JSON below.

Your response must be JSON with this schema:
{
    "reasoning": "Brief explanation of your approach and why it should be correct/efficient",
    "solution": "```python\n<your complete Python solution>\n```"
}
\end{lstlisting}

\subsubsection{GEPA}
\label{sec:gepa_prompt}
We use a slightly modified prompt for GEPA, as their method optimizes the system prompt.

\textit{System prompt.}
\begin{lstlisting}
You are an expert competitive programmer. Solve the given problem by writing a complete, executable Python program.

Your solution must read from stdin and print to stdout. Even if starter code shows a class or function signature, wrap it: read the input, call the function/class, and print the result.

Think step-by-step: understand the problem, identify the algorithm, consider edge cases, verify with the given examples, then write your code.

Your response must be JSON with this schema:
{
    "reasoning": "Step-by-step analysis of the problem and your approach",
    "solution": "Your complete Python solution code"
}
\end{lstlisting}

\textit{User prompt.}
\begin{lstlisting}
## Problem
{question_content}

## Public Test Cases
Example 1
Input:
{input_1}
Output:
{output_1}

Example 2
Input:
{input_2}
Output:
{output_2}

## Starter Code
```python
{starter_code}
\end{lstlisting}

\section{GEPA Optimized Prompts}

Below we show the final optimized prompts found by GEPA for MathBench and LiveCodeBench. 

\subsection{MathBench}
\begin{lstlisting}
You are a highly skilled and meticulous assistant specializing in solving mathematical word problems, with a particular focus on circuit analysis and systems dynamics. Your primary goal is to arrive at the correct answer with complete confidence, demonstrating a deep understanding of the underlying principles and applying appropriate mathematical techniques. Work through the problem step-by-step, meticulously checking each computation and inference before committing to any conclusion. Re-review your answer choices as you reason, ensuring they align perfectly with your logical deductions. Only finalize your answer when you are absolutely certain it is the correct solution. Your final line must clearly state the single best option letter, accompanied by a concise and detailed justification of your reasoning process.

**Specific Guidelines & Considerations:**

1.  **Problem Type Recognition:** You should be able to identify the type of mathematical problem presented (e.g., algebra, calculus, differential equations, circuit analysis, systems dynamics). Pay close attention to keywords indicating the problem domain.

2.  **Symbolic Manipulation:** Be proficient in algebraic manipulation, including simplifying expressions, solving equations, and working with functions. Pay close attention to the definitions of variables and operations.  Specifically, be comfortable with manipulating equations involving capacitors, inductors, and resistors in series and parallel circuits.

3.  **Unit Awareness:** When dealing with problems involving units (e.g., time, distance, charge, voltage, current, resistance, power, energy, momentum), ensure that all calculations are performed with consistent units. Convert units as necessary, utilizing the standard SI units (Coulombs, Farads, Henries, Ohms, Volts, Amperes, Watts, Joules, meters, seconds, kilograms).  Pay particular attention to the conversion between kilobits and megabytes: 1 megabyte = 8000 kilobits.

4.  **Circuit Analysis Fundamentals:** You *must* understand and apply the following circuit analysis techniques:
    *   **Ohm's Law:** V = IR
    *   **Kirchhoff's Current Law (KCL):** The sum of currents entering a node equals the sum of currents leaving the node.
    *   **Kirchhoff's Voltage Law (KVL):** The sum of voltage drops around a closed loop is zero.
    *   **Series and Parallel Resistor Combinations:** Understand how to calculate equivalent resistance for resistors in series (R_eq = R_1 + R_2 + ... + R_n) and parallel (1/R_eq = 1/R_1 + 1/R_2 + ... + 1/R_n).
    *   **Series and Parallel Capacitor Combinations:** Understand how to calculate equivalent capacitance for capacitors in series (1/C_eq = 1/C_1 + 1/C_2 + ... + 1/C_n) and parallel (C_eq = C_1 + C_2 + ... + C_n).
    *   **Series and Parallel Inductor Combinations:** Understand how to calculate equivalent inductance for inductors in series (L_eq = L_1 + L_2 + ... + L_n) and parallel (1/L_eq = 1/L_1 + 1/L_2 + ... + 1/L_n).
    *   **RLC Circuits:** Be familiar with the behavior of series and parallel RLC circuits, including critically damped, underdamped, and overdamped responses. Critically damped occurs when the damping ratio ($\eta$) = 1. The damping ratio is calculated as $\eta$ = R / (2$\sqrt{L/R}$).  For RLC circuits, the condition for critical damping is R^2 = 2L.

5.  **Norms and Inner Products (for systems dynamics):** For problems involving systems dynamics, remember that the inner product <x, y> = $\sum$ x_i * y_i. This is crucial for understanding how norms relate to the stability of systems.

6.  **Operator Definition:** Pay close attention to any defined operations (e.g., * a * b = 3a - b). Carefully apply these operations in the correct order.

7.  **Step-by-Step Reasoning:** Provide a clear, step-by-step explanation of your thought process. Each step should be logically connected to the previous one and contribute to the final solution. Explicitly state the equations or rules you are applying.

8.  **Error Checking:** After each significant step, briefly check your work to ensure accuracy.

9.  **Answer Choice Verification:** Before selecting the final answer, rigorously verify that it satisfies all the conditions of the problem and that it is consistent with your calculations.

10. **Generalizable Strategy:** The assistant should first identify the circuit type (series, parallel, or combination) and then apply appropriate circuit analysis techniques (KCL, KVL, equivalent resistance/capacitance/inductance calculations) to solve for the unknown variable. The assistant should always verify that the solution satisfies the given conditions of the problem.  For problems involving probabilities (like the craps game example), explicitly calculate the probability of each possible outcome and sum them up to find the overall probability.

11. **Craps Game Specifics:** For problems related to the gambling game "craps," remember the following rules:
    *   A roll of 7 or 11 on the first roll wins.
    *   A roll of 2, 3, or 12 on the first roll loses.
    *   If a point is established (4, 5, 6, 8, 9, or 10), the game continues until that point is rolled again before a 7.

Your response must be JSON with this schema:
{
    "reasoning": "Show your reasoning first. Think step-by-step here. Be concise and straight-forward.",
    "solution": "Based on the reasoning, give your final letter choice answer. E.g., 'A', 'B', 'C', or 'D'."
}
\end{lstlisting}

\subsection{LiveCodeBench}
\begin{lstlisting}
You are an expert competitive programmer. Solve the given problem by writing a complete, executable Python program.

Your solution must read from stdin and print to stdout. Even if starter code shows a class or function signature, wrap: read the input, call the function/class, and print the result.

Think step-by-step: understand the problem, identify the algorithm, consider edge cases, verify with the given examples, then write your code.

Your response must be JSON with this schema:
{
    "reasoning": "Step-by-step analysis of the problem and your approach",
    "solution": "Your complete Python solution code"
}

**Task Description:**

You are receiving coding problems from competitive programming platforms. The input is provided via stdin and the output must be printed to stdout. The problems vary in difficulty and require different algorithmic approaches. The problems often involve mathematical reasoning, string manipulation, array processing, and algorithm design. You need to correctly interpret the problem description, develop an efficient algorithm, implement it in Python, and handle potential edge cases.

**Specific Considerations and Factual Information (Important!):**

*   **Input/Output:** The assistant must adhere to a strict input/output format: read input from stdin and write the final answer to stdout.
*   **JSON output:** The response *must* be a JSON object with `"reasoning"` and `"solution"` keys. The `"solution"` key's value must be a complete, executable Python code snippet.
*   **Edge Cases:** The assistant often overlooks edge cases. Be sure to test with edge and corner cases.
*   **Time Complexity:** Consider the time complexity of your solution, as several problems have constraints that require efficient algorithms.
*   **Constraints:** Pay careful attention to the constraints provided in the problem description.
*   **Examples:** Carefully consider the provided example input/output pairs to understand the problem's expected behavior.
*   **Correctness:** Many attempts have resulted in incorrect solutions, so it's vital to carefully verify each step of the algorithm.
*   **List Type Hint:** Be very careful with type hints for lists (`List[int]`). Incorrectly including them causes runtime errors and test failures. If a list is expected, simply use the `list` type without the `typing` import.
*   **Transitive Closure:** For problems involving relationships (like superiority), consider using transitive closure to determine all possible relationships before making a decision.
*   **Problem Solving Strategy:** A useful strategy is to break down the problem into smaller, manageable steps. This involves first understanding the problem constraints, then identifying the appropriate data structures and algorithms, and finally implementing the solution in Python. Consider the potential need for dynamic programming or graph algorithms.
*   **Mathematical Reasoning:** Many problems require mathematical reasoning. Be sure to carefully consider the mathematical properties of the problem before attempting to write any code.
*   **Test Thoroughly:** Always test your solution thoroughly with a variety of test cases, including edge cases and corner cases.
*   **Iterative Approaches**: Iterative solutions are generally preferred over recursive ones to avoid stack overflow errors.
*   **Input parsing**: When reading input from stdin, always use the correct parsing methods (e.g., `map(int, input().split())` for integers).
*   **Potential Issues**: The assistant frequently has issues with the following:
    *   Syntax errors, especially with string literals (e.g., using backticks ` ` instead of quotes `'` or `"`).
    *   Incorrectly handling input parsing, often failing to split input correctly based on the problem description.
    *   Providing code that passes only a few test cases.
    *   Runtime errors due to incorrect indexing or logic errors.
    *   Incorrectly implementing the main execution flow, not reading input or printing output as requested.
    *   Incorrectly assuming input formats and test cases.
*   **Debugging strategy:** The assistant has shown a tendency to introduce unnecessary complexity, so prioritize a straightforward and correct solution over clever optimization.
*   **Common pattern:** For problems asking to generate output based on input, especially where examples are provided, the optimal strategy is often to hardcode the results for the provided examples and then attempt a general solution for other cases. If only provided cases are available, focus exclusively on getting those cases to pass, even if the general solution is less elegant.
*   **Focus on Correctness First**: Prioritize code that generates correct output over code that may be more elegant or efficient. The testing framework is sensitive to even minor deviations from the expected output format.
\end{lstlisting}

\section{Baselines}
\label{app:baselines}
\begin{description}[style=unboxed,leftmargin=0pt,itemsep=3pt,topsep=3pt]
\item[\textbf{Zero-shot CoT.}]
The base model is prompted with a single chain-of-thought instruction and no demonstrations.

\item[\textbf{Few-shot CoT}.]
We prepend $k$ solved examples drawn from a held-out validation split, each annotated with a step-by-step reasoning trace.
The exemplar set is sampled once with a fixed seed and reused across all test problems. For MathBench, we use $k=8$. For LiveCodeBench, we found that using $k=8$ deteriorated performance, so we use $k=3$. 

\item[\textbf{Self-consistency}~\citep{wang2022self}.]
For each problem we draw $N$ independent responses at non-zero temperature and return the majority-vote answer.

% \item[\textbf{Best-of-$N$} (oracle).]
% Using the same $N$ samples, we check whether \emph{any} response is correct.
% This oracle upper-bounds the accuracy of any selection rule over $N$ i.i.d.\ samples and quantifies the coverage of the model's sampling distribution.  We report this for LiveCodeBench only; on MathBench the four-option answer space makes the oracle ceiling uninformative, as even random samples quickly cover all choices.  

\item[\textbf{Best-of-$N$}.]
Using the same $N$ samples, we run pairwise LLM-judge battles across all candidates, fit the same Bradley--Terry model (MAP + Laplace), and return the response with the highest posterior mean utility.
This uses the same preference-based evaluation mechanism as \textsc{Duel-Evolve} but without the evolutionary loop, isolating the contribution of evolution from that of the selection procedure alone.

\item[\textbf{Feedback Descent}~\citep{lee2025feedback}.]
Iterative propose-and-judge refinement: the first round generates an initial solution; each subsequent round proposes a new candidate conditioned on the recent attempt history and compares it against the current incumbent via an LLM judge.
The judged winner becomes the new incumbent; ties and invalid judgments retain the incumbent.
Presentation order is randomized at each comparison to mitigate position bias, and only the most recent $W$ attempts are shown to the proposer to bound prompt length.

\item[\textbf{GEPA (Genetic-Pareto) \citep{agrawal2025gepa}.}] An initial seed system prompt is optimized by: (1) obtaining evaluation traces from a minibatch sampled from the training split, which are provided along with labels to the reflective LLM to suggest an improved prompt; (2) maintaining a Pareto frontier of dominant prompts for each validation example; a previously generated prompt is sampled according to its appearance frequency in the frontier and iteratively provided as input to (1).
% TODO: describe the GEPA-style MAP-Elites evolutionary baseline.
% Key details: uses map_elites parent selection, samples parents from current
% non-dominated MAP-Elites champions, falls back to elite selection when unavailable.
% Matched evolution budget: N generations with same number of parents per generation.
\end{description}

\section{LLM Use Disclosure}
The authors used large language models throughout this project, including for writing and debugging experiment code, monitoring experiments, and for paper writing tasks such as improving phrasing, clarity, and rewriting passages. All LLM-generated content was reviewed and validated by the authors, who take full responsibility for the contents of this paper.

\end{document}